\documentclass[conference,10pt]{IEEEtran}
\IEEEoverridecommandlockouts
\usepackage{cite}
\usepackage[pdftex]{graphicx}
\usepackage{times,amsmath}
\usepackage{color}
\usepackage{theorem}
\usepackage{amssymb}
\usepackage{subfigure}
\usepackage{hyperref}
\usepackage[ruled,vlined]{algorithm2e}
\usepackage{hyphenat}
\usepackage{tikz,chemarrow}
\usepackage[framemethod=tikz]{mdframed}
\usepackage[normalem]{ulem}

\input{my_symbol.sty}

\newtheorem{mytheorem}{Theorem}

\def\BibTeX{{\rm B\kern-.05em{\sc i\kern-.025em b}\kern-.08em
    T\kern-.1667em\lower.7ex\hbox{E}\kern-.125emX}}
    
\begin{document}

\title{Online Graph Learning under Smoothness Priors}
\author{Seyed Saman Saboksayr\IEEEauthorrefmark{1}, Gonzalo Mateos\IEEEauthorrefmark{1}\IEEEauthorrefmark{2} and Mujdat Cetin\IEEEauthorrefmark{1}\IEEEauthorrefmark{2}\\ 
\IEEEauthorrefmark{1}Dept. of Electrical and Computer Engineering, University of Rochester, Rochester, NY, USA\\
\IEEEauthorrefmark{2}Goergen Institute for Data Science, University of Rochester, Rochester, NY, USA
\thanks{Work in this paper was supported by the NSF awards CCF-1750428, CCF-1934962 and ECCS-1809356. Author emails: ssaboksa@ur.rochester.edu, gmateosb@ece.rochester.edu, and mujdat.cetin@rochester.edu.}}

% \author{\IEEEauthorblockN{Seyed Saman Saboksayr}
% \IEEEauthorblockA{\textit{Dept. of Electrical and Computer Eng.} \\
% \textit{Univ. of Rochester}\\
% Rochester, USA\\
% ssaboksa@ur.rochester.edu}
% \and
% \IEEEauthorblockN{Gonzalo Mateos}
% \IEEEauthorblockA{\textit{Dept. of Electrical and Computer Eng.} \\
% \textit{Univ. of Rochester}\\
% Rochester, USA\\
% gmateosb@ece.rochester.edu}
% \and
% \IEEEauthorblockN{Mujdat Cetin}
% \IEEEauthorblockA{\textit{Dept. of Electrical and Computer Eng.} \\
% \textit{Univ. of Rochester}\\
% Rochester, USA\\
% mujdat.cetin@rochester.edu}
% }

\maketitle

\begin{abstract}
The growing success of graph signal processing (GSP) approaches relies heavily on prior identification of a graph over which network data admit certain regularity. However, adaptation to increasingly dynamic environments as well as demands for real-time processing of streaming data pose major challenges to this end. In this context, we develop novel algorithms for online network topology inference given streaming observations assumed to be smooth on the sought graph. Unlike existing batch algorithms, our goal is to track the (possibly) time-varying network topology while maintaining the memory and computational costs in check by processing graph signals sequentially-in-time. To recover the graph in an online fashion, we leverage proximal gradient (PG) methods to solve a judicious smoothness-regularized, time-varying optimization problem. Under mild technical conditions, we establish that the online graph learning algorithm converges to within a neighborhood of (i.e., it tracks) the optimal time-varying batch solution. Computer simulations using both synthetic and real financial market data illustrate the effectiveness of the proposed algorithm in adapting to streaming signals to track slowly-varying network connectivity.
\end{abstract}
\begin{IEEEkeywords}
Graph learning, graph signal processing, online optimization, smooth signals, network topology inference.
\end{IEEEkeywords}
%
%%%%%%%%%%%%%%%%%%%%%%%%%%%%%%%%%%%%%%%%%%%%%%%%%%%%%%%%%%%%%%%%%%%%%

\section{Introduction}\label{S:Introduction}

%%%%%%%%%%%%%%%%%%%%%%%%%%%%%%%%%%%%%%%%%%%%%%%%%%%%%%%%%%%%%%%%%%%%%

Making sense of relational datasets from a network-centric perspective is essential to obtain actionable information in various fields of science and engineering. Graph signal processing (GSP) has played a key role to that end, underscoring the value of graphs as models of complex signals with irregular structures~\cite{ortega18,kolaczyk09}. However, said graphs are often not readily available and a first crucial step is to use nodal observations (i.e., measurements of graph signals) to identify the network structure, or, a useful graph model that facilitates signal representations and downstream learning tasks; see~\cite{mateos19,dong2019learning} for tutorials on recent network topology inference advances. Acknowledging that many of these networks are also dynamic (e.g., in applications involving financial markets), there is a growing need to develop online graph learning algorithms that can process network data streams in an efficient manner~\cite{giannakis18}.

Given a set of graph signal observations, the network topology inference problem is to find the graph (represented through e.g., an adjacency or Laplacian matrix) that is optimal in some sense. The optimality criterion is usually dictated by the adopted network-dependent model for the measurements, often augmented by structural (e.g., edge sparsity) priors motivated by physical characteristics to effect statistical regularization or favor interpretability. Network data models are often given by probabilistic priors such as Gaussian Markov random fields (GMRFs), where the graph learning problem becomes one of graphical model (here covariance) selection~\cite[Ch. 7]{kolaczyk09}. Other models are specified via signal parsimony-related properties with respect to the underlying graph, including stationarity~\cite{segarra2016topoidTSP16,rasoul20} and smoothness (i.e., bandlimitedness, linked to GMRF selection under Laplacian constraints)~\cite{kalofolias16,dong16,kalofolias2019iclr}.
% \cite{friedman08, pasdeloup2016inferenceTSIPN16, bars19, sardellitti19, berger2020graphlearning}

\noindent \textbf{Contributions in context of related prior work.} In this paper, we develop an online algorithmic framework to estimate (possibly dynamic) graphs under smoothness priors. Many real-world signals are smooth over judicious networks, including temperature recordings~\cite{chepuri17}, movie ratings, and natural images~\cite{kalofolias17}, to name a few. Exploitation of this cardinal property is at the heart of several graph-based statistical learning tasks including nearest-neighbor prediction (also known as graph smoothing), denoising, semi-supervised learning, and spectral clustering~\cite{ortega18,kolaczyk09}. Revisiting the general graph learning framework in~\cite{kalofolias16,kalofolias2019iclr} -- but with streaming data -- we develop online proximal-gradient (PG) iterations to solve the resulting smoothness-regularized, time-varying optimization problem. There are noteworthy recent works on time-varying network topology inference from observations of smooth signals~\cite{cardoso20,kalofolias17,koki20}. Unlike our online algorithm, those in~\cite{cardoso20,kalofolias17,koki20} operate in batch mode, they are non-recursive, and hence their computational complexity and memory storage grow linearly with time. Borrowing techniques from~\cite{madden19}, we establish that the online PG algorithm converges to within a neighborhood of (i.e., it tracks) the optimal time-varying batch solution. To the best of our knowledge, this is the first work that addresses the problem of online graph learning from streaming smooth signals with quantifiable performance guarantees. Different from our signal smoothness assumption, the online graph learning scheme in~\cite{vlaski2018online} uses observations from a Laplacian-based, continuous-time graph process and \cite{rasoul20} relies on stationarity. Numerical tests using both synthetic and real financial market data corroborate the efficiency and effectiveness of the proposed PG algorithm in adapting to streaming signals and tracking changes in the sought (slowly-varying) dynamic network.

%%%%%%%%%%%%%%%%%%%%%%%%%%%%%%%%%%%%%%%%%%%%%%%%%%%%%%%%%%%%%%%%%%%%%

\section{Graph Signal Processing Preliminaries}
\label{S:Preliminaries}

%%%%%%%%%%%%%%%%%%%%%%%%%%%%%%%%%%%%%%%%%%%%%%%%%%%%%%%%%%%%%%%%%%%%%

Consider a weighted, undirected graph $\ccalG \left ( \ccalV, \ccalE, \bbW \right)$, where $\ccalV = \left \{ 1,\dots,N \right \}$ is the set of vertices, $\ccalE \subseteq \ccalV \times \ccalV$ denotes the set of edges, and $\bbW \in \reals^{N \times N}_{+}$ is the symmetric adjacency matrix. In the absence of connection [i.e., $\left (i,j \right ) \nsubseteq \ccalE$] one has $W_{ij}=0$. Graph $\ccalG$ is devoid of self-loops, which implies $W_{ii}=0$, $\forall i\in\ccalV$. We will henceforth assume nodal degrees $\bbd:=\bbW\mathbf{1}$ are uniformly lower bounded away from zero, i.e., $\bbd\succeq d_{\min}\mathbf{1}$ (entry-wise inequality) for some prescribed $d_{\min}>0$. Else if degrees become arbitrarily small, it is prudent to apply a threshold and remove the loosely connected nodes from $\ccalG$. We study graph signals $\bbx =\left[x_1,\dots,x_N\right]^{\top}\in\reals^N$ defined on $\ccalG$, where $x_i$ is the signal value at node $i \in \ccalV$. Directed graphs could be useful models~\cite{marques20}, but are beyond the scope of this paper. Complex-valued signals can be accommodated as well~\cite{ortega18}.\vspace{2pt}

\noindent{\textbf{Signal smoothness with respect to $\ccalG$.}} The adjacency matrix $\bbW$ encodes the graph's topology. Beyond $\bbW$, advances in spectral graph theory often motivate choosing the  combinatorial graph Laplacian $\bbL := \diag \left( \bbd \right) - \bbW$. % as well as its various degree-normalized counterparts. 
In particular, $\bbL$ plays a central role in defining a graph Fourier transform (GFT)~\cite{ortega18} as well as a measure of signal variability with respect to $\ccalG$~\cite{zhou04}. Focusing on the latter, the  total variation (TV) of the graph signal $\bbx$ with respect to the Laplacian $\bbL$ (also known as Dirichlet energy) is defined as the following quadratic form:
\begin{equation}\label{eq5}
    \textrm{TV}(\bbx):=\bbx^{\top}\bbL \bbx
    = \frac{1}{2}\sum_{i \neq j} W_{ij} \left( x_i - x_j \right)^2.
\end{equation}
The $\textrm{TV}(\bbx)$ is a smoothness measure, quantifying how much the graph signal $\bbx$ changes with respect to $\ccalG$'s topology. Smaller values of $\textrm{TV}(\bbx)$ are indicative of limited signal variability, with $\textrm{TV}(\alpha\mathbf{1})=0$ as an extreme. More germane to the graph-learning theme of this paper is to use smoothness as the criterion to construct graphs on which network data admit certain regularity, the subject dealt with next.

%%%%%%%%%%%%%%%%%%%%%%%%%%%%%%%%%%%%%%%%%%%%%%%%%%%%%%%%%%%%%%%%%%%%%

\section{Graph learning from smooth signals}\label{S:Batch}

%%%%%%%%%%%%%%%%%%%%%%%%%%%%%%%%%%%%%%%%%%%%%%%%%%%%%%%%%%%%%%%%%%%%%

Consider the following network topology identification problem. Given a set $\ccalX:=\{\bbx_t\}_{t=1}^T$ of possibly noisy graph signal observations acquired at time $t$, the goal is to learn an undirected graph $\ccalG(\ccalV,\ccalE, \bbW)$ with $|\ccalV|=N$ nodes such that the observations in $\ccalX$ are smooth on $\ccalG$. The graph can be dynamic with a slowly time-varying adjacency matrix $\bbW_t$, $t=1,2,\ldots$ (see Section \ref{S:Online}), but for now we omit any form of temporal dependency to simplify exposition. In this section, we briefly review the general graph learning framework proposed in~\cite{kalofolias16,kalofolias2019iclr}, that we build on in the rest of the paper.\vspace{2pt}

\noindent{\textbf{Graph learning under smoothness priors.}} Given $\ccalX$ one can form the data matrix $\bbX=[\bbx_1,\ldots,\bbx_T]\in \reals^{N\times T}$, and let $\bar{\bbx}_i^{\top}\in\reals^{1\times T}$ denote its $i$-th row collecting those $T$ measurements at vertex $i$. The neat idea in~\cite{kalofolias16} is to establish a link between smoothness and sparsity, namely
\begin{equation}\label{E:smooth_sparse}
	\sum_{t=1}^T\textrm{TV}(\bbx_t)=\textrm{trace}(\bbX^{\top}\bbL\bbX)=\frac{1}{2}\|\bbW\circ\bbZ\|_1,
\end{equation}
where $\circ$ stands for the Hadamard (element-wise) product and the Euclidean-distance matrix $\bbZ\in\reals_{+}^{N\times N}$ has entries $Z_{ij}:=\|\bar{\bbx}_i-\bar{\bbx}_j\|^2$, $i,j\in\ccalV$. The intuition is that when the given distances in $\bbZ$ come from a smooth manifold, the corresponding graph has a sparse edge set, with preference given to edges $(i,j)$ associated with smaller distances $Z_{ij}$. 

Leveraging \eqref{E:smooth_sparse} a general graph-learning framework was put forth in~\cite{kalofolias16}, which advocates solving the convex smoothness-regularized inverse problem
\begin{align}\label{eq:kalofolias}
	\min_{\bbW}&{}\:\|\bbW\circ\bbZ\|_1+g(\bbW)\\
	\textrm{ s. t. } &{} \quad\textrm{diag}(\bbW)=\mathbf{0},\: W_{ij}=W_{ji}\geq 0, \:i\neq j.\nonumber
\end{align}
The convex objective function $g(\bbW)$ encodes additional structural properties of $\ccalG$. Several choices for $g(\bbW)$ have been proposed to, e.g., recover graphs based on the Gaussian kernel~\cite{friedman01}, accommodate time-varying graphs~\cite{kalofolias17}, or to scale other related graph learning algorithms~\cite{dong16}.
Identity \eqref{E:smooth_sparse} offers a favorable way of formulating the inverse problem \eqref{eq:kalofolias}, because the space of adjacency matrices can be described via simpler (meaning entry-wise decoupled) constraints relative to its Laplacian counterpart. As a result, \eqref{eq:kalofolias} can be solved efficiently with complexity $\ccalO(N^2)$ per iteration, by leveraging provably-convergent primal-dual solvers. Next, we present a different optimization approach based on PG methods~\cite{boyd14}, which offers an (equally) efficient alternative in the batch setting while it lends itself naturally to online operation.\vspace{2pt}

\noindent{\textbf{Batch proximal gradient algorithm.}} To make \eqref{eq:kalofolias} amenable to the PG method, recall first that the adjacency matrix $\bbW$ is symmetric with diagonal elements equal to zero. Thus, the independent decision variables are effectively the upper-triangular elements $[\bbW]_{ij}$, $j>i$, which we collect in the vector $\bbw \in \reals_{+}^{N(N-1)/2}$. Second, it will prove convenient to enforce the non-negativity constraints via a penalty function augmenting the original objective. Just like~\cite{kalofolias16} we add an indicator function $\ind{\bbw\succeq\mathbf{0}}=0$ if $\bbw\succeq \mathbf{0}$, and $\ind{\bbw\succeq \mathbf{0}}=\infty$ otherwise. Given these definitions we recast the objective in \eqref{eq:kalofolias} as the function $F(\bbw)$ of a vector variable, and write the equivalent composite, non-smooth optimization problem
\begin{align}\label{eq:kalofolias_vec}
	\min_{\bbw}&{}\:F\left(\bbw\right) :=\underbrace{\mbI\left\{\bbw\succeq\mathbf{0}\right\} + 2\bbw^{\top}\bbz}_{h(\bbw)}+ g(\bbw), 
\end{align}
where $\bbz$ is a vector containing the upper-triangular entries of $\bbZ$, and $\bbS\in\{0,1\}^{N\times N(N-1)/2}$ is such that $\bbd=\bbW\mathbf{1}=\bbS\bbw$. 

A useful choice is $g(\bbw)=\beta 2\| \bbw\|^2-\alpha \bbone^{\top} \log \left( \bbS\bbw \right)$, where $\alpha,\beta>0$ are tuning parameters~\cite{kalofolias16}. The logarithmic barrier on the nodal degree sequence $\bbS\bbw$ precludes the trivial solution $\bbw=\mathbf{0}$. Moreover, it ensures the estimated graph is devoid of isolated vertices. The $\ell_2$-norm regularization on the adjacency matrix $\bbw$ controls the graphs' edge sparsity pattern by penalizing larger edge weights (the sparsest graph is obtained for $\beta=0$). The gradient of said $g$ has the form
% simple
\begin{equation} \label{eq:grad_g}
	\nabla g(\bbw) = 4\beta\bbw-\alpha \bbS^{\top}\left(\frac{\bbone}{\bbS\bbw}\right),
\end{equation}
where $\mathbf{1}/\bbS\bbw$ stands for element-wise division. Moreover,  $\nabla g$ is a Lipschitz-continuous function with constant $\eta=\left ( 4\beta + \frac{2\alpha (N-1)}{d_{\min}^2} \right )$; see~\cite{saboksayr20} for the proof that is omitted here due to lack of space.

For constant step size $\mu < \frac{2}{\eta}$, the PG iterations to solve the batch graph learning problem \eqref{eq:kalofolias_vec} are given by (henceforth $k=0,1,2,\ldots$ denote iterations)
\begin{equation} \label{eq:batch_prox}
	\bbw_{k+1} = \textbf{prox}_{\mu h}\left(\bbw_{k} - \mu \nabla g(\bbw_{k}) \right),
\end{equation}
where the proximal operator of $h$ in \eqref{eq:kalofolias_vec} is 
\begin{equation}\label{eq:prox_h}
	\textbf{prox}_{\mu h} (\bbw)=\max\left(\mathbf{0}, \bbw - 2\mu\bbz \right).
\end{equation}
The non-negative soft-thresholding operator in \eqref{eq:prox_h} sets to zero all edge weights in $\bbw$ that fall below the data-dependent thresholds in vector $2\mu\bbz$ (the $\max$ operator is applied entry-wise). Inspection of \eqref{eq:batch_prox} shows that graph estimate refinements are generated via the composition of a gradient-descent step and a proximal operator. 

All in all, \eqref{eq:batch_prox} scales well to large graphs with thousands of nodes and it is competitive with the state-of-the-art primal-dual solver  in~\cite{kalofolias16}. In terms of convergence, as $k\to\infty$ the sequence of iterates \eqref{eq:batch_prox} provably approaches a minimizer of the composite cost $F$ in \eqref{eq:kalofolias_vec}; see e.g.,~\cite{boyd14} for the technical details. Moreover, the worst-case convergence rate of PG algorithms is well documented (namely $\ccalO(1/\epsilon)$ iteration complexity to return a $\epsilon$-optimal solution measured in terms of $F$ values), and can be boosted to $\ccalO(1/\sqrt{\epsilon})$ via Nesterov-type acceleration techniques. Building on recent advances in time-varying convex optimization~\cite{madden19}, our main contribution is to develop novel online PG algorithms for time-varying graphs in the previously unexplored streaming setting.
%\cite{dallanese20,madden19}

%%%%%%%%%%%%%%%%%%%%%%%%%%%%%%%%%%%%%%%%%%%%%%%%%%%%%%%%%%%%%%%%%%%%%

\section{Online Graph Learning}\label{S:Online}

%%%%%%%%%%%%%%%%%%%%%%%%%%%%%%%%%%%%%%%%%%%%%%%%%%%%%%%%%%%%%%%%%%%%%

We switch gears to online estimation of $\bbW$ (or even tracking $\bbW_t$
in a dynamic setting) from streaming data $\{\bbx_1,\ldots,\bbx_t,\bbx_{t+1},\ldots\}$.
To this end, a viable approach is to solve at each time instant $t = 1, 2,\dots$, the composite, time-varying optimization problem [cf. \eqref{eq:kalofolias_vec}]
\begin{multline}\label{eq:online}
    \bbw_{t}^{\star} \in \argmin_{\bbw} F_t\left(\bbw\right):= \overbrace{\mbI\left\{\bbw\succeq \mathbf{0}\right\} + 2\bbw^{\top}\bbz_{1:t}}^{h_t(\bbw)}\\
    \underbrace{-\alpha \bbone^{\top} \log \left( \bbS\bbw \right) + \beta 2\| \bbw\|^2.}_{g(\bbw)}
\end{multline}
In writing $\bbz_{1:t}$ we make explicit that the Euclidean-distance matrix is computed using all signals acquired by time $t$. As data come in, the edge-wise $\ell_1$-norm weights  will fluctuate explaining the time dependence of $F_t(\bbw)$ through its non-smooth component $h_t$.\vspace{2pt}

\noindent{\textbf{Online algorithm construction.}} A naive sequential estimation approach consists of solving \eqref{eq:online} repeatedly using the batch PG algorithm in Section \ref{S:Batch}. However (pseudo) real-time operation in delay-sensitive applications may not tolerate running multiple inner PG iterations per time interval,
so that convergence to $\bbw_{t}^{\star}$ is attained for each $t$. For time-varying graphs it may not be even prudent to obtain $\bbw_{t}^{\star}$ with high precision (hence incurring high delay and unnecessary computational cost), since at time $t+1$ a new datum arrives and the solution $\bbw_{t+1}^{\star}$ may be substantially off the prior estimate. These reasons motivate devising an efficient online and recursive algorithm to solve the time-varying optimization problem \eqref{eq:online}.

%To this end, we build on the online PG methods in~\cite{madden19}. 
Our approach entails two steps per time instant $t=1,2,\ldots$. First, we recursively update the upper-triangular entries $\bbz_{1:t}=\bbz_{1:t-1}+\bbz_t$ of the Euclidean-distance matrix via an exponential moving average (EMA), namely
%  algorithm construction \cite{dallanese20,madden19}
\begin{equation}\label{eq25}
    \barbz_{t} = (1-\gamma)\barbz_{t-1}+\gamma\bbz_{t}. 
\end{equation}
The constant $\gamma\in (0,1)$ is a discount factor, which downweighs past data to facilitate tracking dynamic graphs in non-stationary environments. The larger the constant $\gamma$, the faster EMA discounts past observations. Second, we run
a single iteration of the batch graph learning algorithm developed in Section \ref{S:Batch} to update $\bbw_{t+1}$, namely
\begin{equation} \label{eq:onlinr_prox}
	\bbw_{t+1} = \textbf{prox}_{\mu_t h_t}\left(\bbw_{t} - \mu_t \nabla g(\bbw_{t}) \right),
\end{equation}
The resulting iterations are tabulated as Algorithm \ref{A:online}.

%%%%%%%%%%%%%%%%%%%%%%%%%%%%%%%%%%%%%%%%%%%%%%%
%%           PG ALGORITHM                  %
%%%%%%%%%%%%%%%%%%%%%%%%%%%%%%%%%%%%%%%%%%%%%%%

\begin{algorithm}[t]\label{A:online}
	\SetAlgoLined
	\textbf{Input} parameters $\alpha, \beta, \gamma,$ stream $\bbz_1,\bbz_2,\ldots$, initial $\bbw_{1},\barbz_{0}$. \\
	\For{$t=1,2,\dots,$}{
		Compute $\nabla g(\bbw_{t}) = 4\beta\bbw_{t}-\alpha \bbS^{\top}\left(\frac{\mathbf{1}}{\bbS\bbw_{t}}\right)$.\\
		Update $\barbz_{t} = (1-\gamma)\barbz_{t-1}+\gamma\bbz_{t}$.\\
		Update $\mu_{t} =\left ( 4\beta + \frac{2\alpha (N-1)}{\min(\bbS\bbw_{t})^2} \right )^{-1}$.\\
		Update $\bbw_{t+1} =\max\left(\mathbf{0}, \bbw_{t} -\mu_t\nabla g(\bbw_{t}) - 2\mu_t\barbz_{t}\right)$.
		}
	\caption{Online graph learning via PG}
\end{algorithm}

%%%%%%%%%%%%%%%%%%%%%%%%%%%%%%%%%%%%%%%%%%%%%%%

The computational complexity is dominated by the gradient evaluation in \eqref{eq:grad_g}, incurring a cost of $\ccalO(N^2)$ per instant $t$ due to scaling and additions of vectors of length $N(N-1)/2$. For sparse graphs, the iterates $\bbw_{t}$ tend to become (and remain) quite sparse at early stages of the algorithm by virtue of the soft-thresholding operations (a sparse initialization $\bbw_{1}$ is preferable). It is thus possible to reduce the complexity further if Algorithm \ref{A:online} is implemented carefully using sparse vector operations. Unlike recent approaches that learn dynamic graphs from the observation of smooth signals~\cite{cardoso20,kalofolias17,koki20}, Algorithm 1’s memory storage requirement and computational cost per data sample $\bbx_t$ does not grow with $t$.\vspace{2pt} 

\noindent \textbf{Convergence analysis.} Here we establish that Algorithm 1 can closely track the sequence of minimizers $\bbw_{t}^{\star}$ for large enough $t$; see also the simulations in Section \ref{S:Simulations}. Noting that $g$ is $4\beta$-strongly convex and its gradient $\eta$-Lipschitz continuous, we can derive bounds for the tracking error $\| \bbw_{t} - \bbw_{t}^{\star} \|_{F}$. To this end, let us define $v_t := \|\bbw_{t+1}^{\star} - \bbw_{t}^{\star} \|_{F}$ to quantify the temporal variability of the optimal solution of \eqref{eq:online}. We have the following (non-asymptotic) performance guarantee for Algorithm 1, adapted from~\cite[Theorem~1]{madden19}.
\begin{mytheorem}\label{th:}
For all $t\geq 1$, the sequence of iterates $\bbw_{t}$ generated by Algorithm \ref{A:online} satisfies:
 \begin{equation}\label{eq.th1}
        \| \bbw_{t} - \bbw_{t}^{\star} \|_{F} \leq \tdL_{t-1}\left(\| \bbw_{0} - \bbw_{0}^{\star} \|_{F} + \sum_{\tau = 0}^{t-1}\frac{v_{\tau}}{\tdL_{\tau}} \right),
\end{equation}
where $L_{t} = \max \left\{ |1-4\mu_t\beta|,|1-\mu_t\eta_t| \right\}$, $\tdL_{t}=\prod_{\tau=1}^{t}L_{\tau}$. Moreover, for the sequence of objective values we can write $F_t(\bbw_{t}) - F_t(\bbw_{t}^{\star})\leq \frac{\eta_t}{2}\| \bbw_{t} - \bbw_{t}^{\star} \|_{F}$; see~\cite[Theorem~10.29]{beck18}.
\end{mytheorem}
To gain further insights let us define $\hhatL_{t} := \max_{\tau=1,\dots,t} L_{\tau}$, $\hhatv_{t} := \max_{\tau=1,\dots,t} v_{\tau}$. The sum of the geometric series in the right-hand side of \eqref{eq.th1} can be simplified to
\begin{equation*}%\label{eq.th2}
     \| \bbw_{t} - \bbw_{t}^{\star} \|_{F} \leq \left(\hhatL_{t-1} \right)^t \| \bbw_{0} - \bbw_{0}^{\star} \|_{F} + \frac{\hhatv_{t}}{1 - \hhatL_{t-1}}.
\end{equation*}
Accordingly, $\hhatL_{t} = (\eta_t - 4\beta)/\eta_{t} < 1$ since $\mu_{t} = \eta_{t}^{-1}$. Therefore, $(\hhatL_{t-1})^t \to 0$ and Algorithm \ref{A:online} converges to the vicinity of the optimal solution with a misadjustment $\hhatv_{t}/(1 - \hhatL_{t-1})$. It follows that the tracking error increases with $\hhatv_{t}$ (rapidly-varying graphs are more challenging) and also if the problem is badly conditioned (i.e., $\beta\to 0$ in which case $\hhatL_{t}\to 1$).
% ; see~\cite{saboksayr20} for more insights.

%%%%%%%%%%%%%%%%%%%%%%%%%%%%%%%%%%%%%%%%%%%%%%%%%%%%%%%%%%%%%%%%%%%%%

\section{Numerical Results}\label{S:Simulations}

%%%%%%%%%%%%%%%%%%%%%%%%%%%%%%%%%%%%%%%%%%%%%%%%%%%%%%%%%%%%%%%%%%%%%

Here we test Algorithm \ref{A:online} on synthetic and real-world financial market data. Throughout, we perform a grid search to determine the best regularization parameters $\alpha,\beta$.\vspace{2pt}

\noindent \textbf{Synthetic data.} To assess the performance of the proposed online graph learning algorithm, we test it on simulated streaming data. We generate a piecewise-constant sequence of two random Erd\H{o}s-R\'enyi graphs (edge formation probability $p=0.15$) with $N=50$ nodes. The initial graph switches after $t=4000$ time samples. The final graph is obtained from the initial one by redrawing $40$ percent of its edges. For each time instant $t=1,\ldots,8000$, we simulate i.i.d. smooth signals with respect to the time-varying underlying graph. The signals are drawn from a Gaussian distribution $\bbx_t \sim \ccalN\left( \bbzero, \bbL_t^{\dag}+\sigma_{e}^2 \bbI_N \right)$, where $\sigma_{e}$ represents the noise level; see e.g.,~\cite{dong16}. Results are averaged over $10$ independent Monte Carlo trials. The simulation is repeated for graphs with $N = 100$ nodes as well. As a baseline, we compute the time-varying solution $F_t$ in \eqref{eq:online} by running the batch PG algorithm until convergence.% for each $500$ time steps. 
% in each of the temporal segments

\begin{figure}[t]
    \centering
   % \centerline{\begin{minipage}[c]{\linewidth}
    %\centering
    \includegraphics[width=0.85\linewidth]{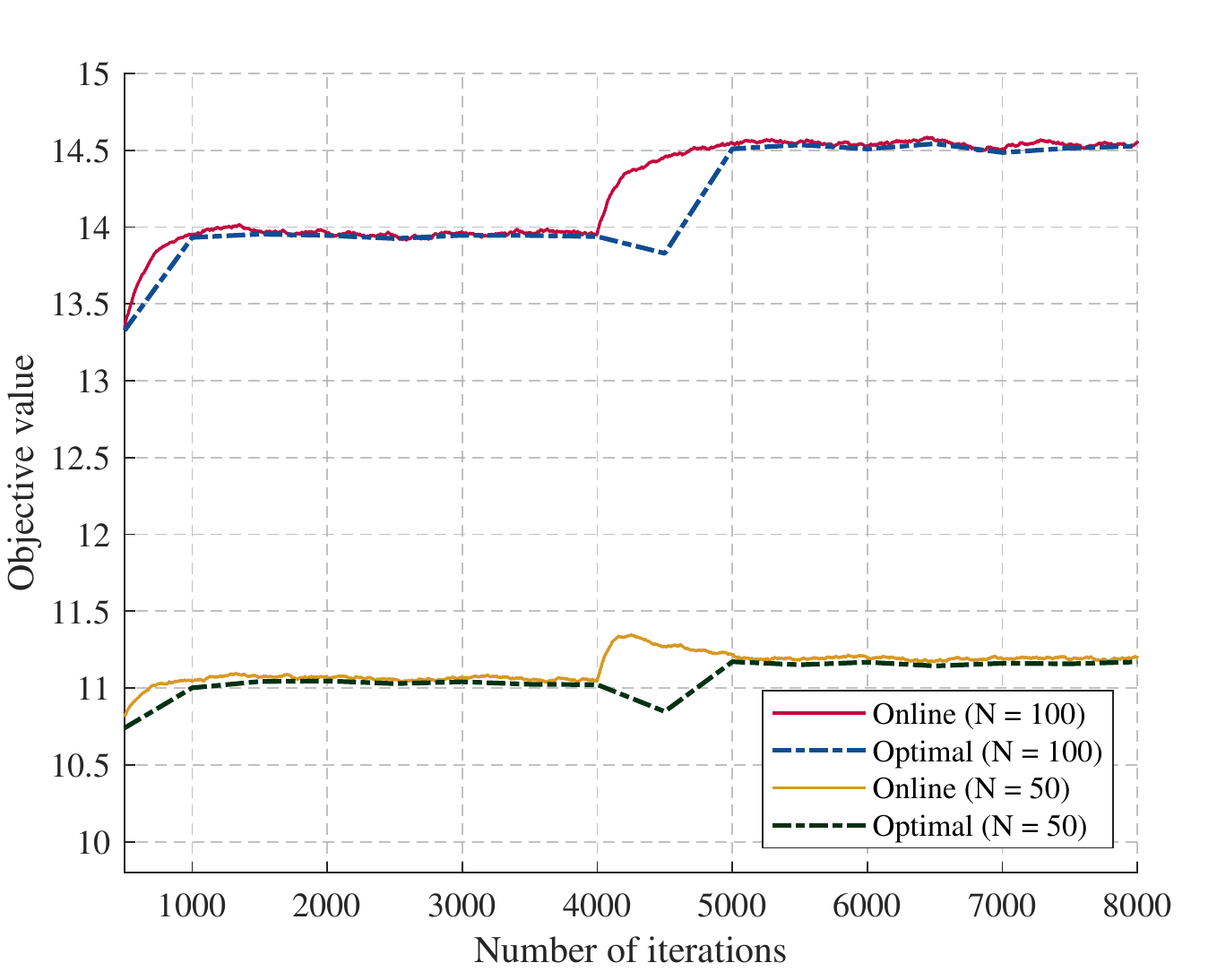}
    %\end{minipage}}
    %\centerline{\begin{minipage}[c]{\linewidth}
    %\centering
    \includegraphics[width=0.85\linewidth]{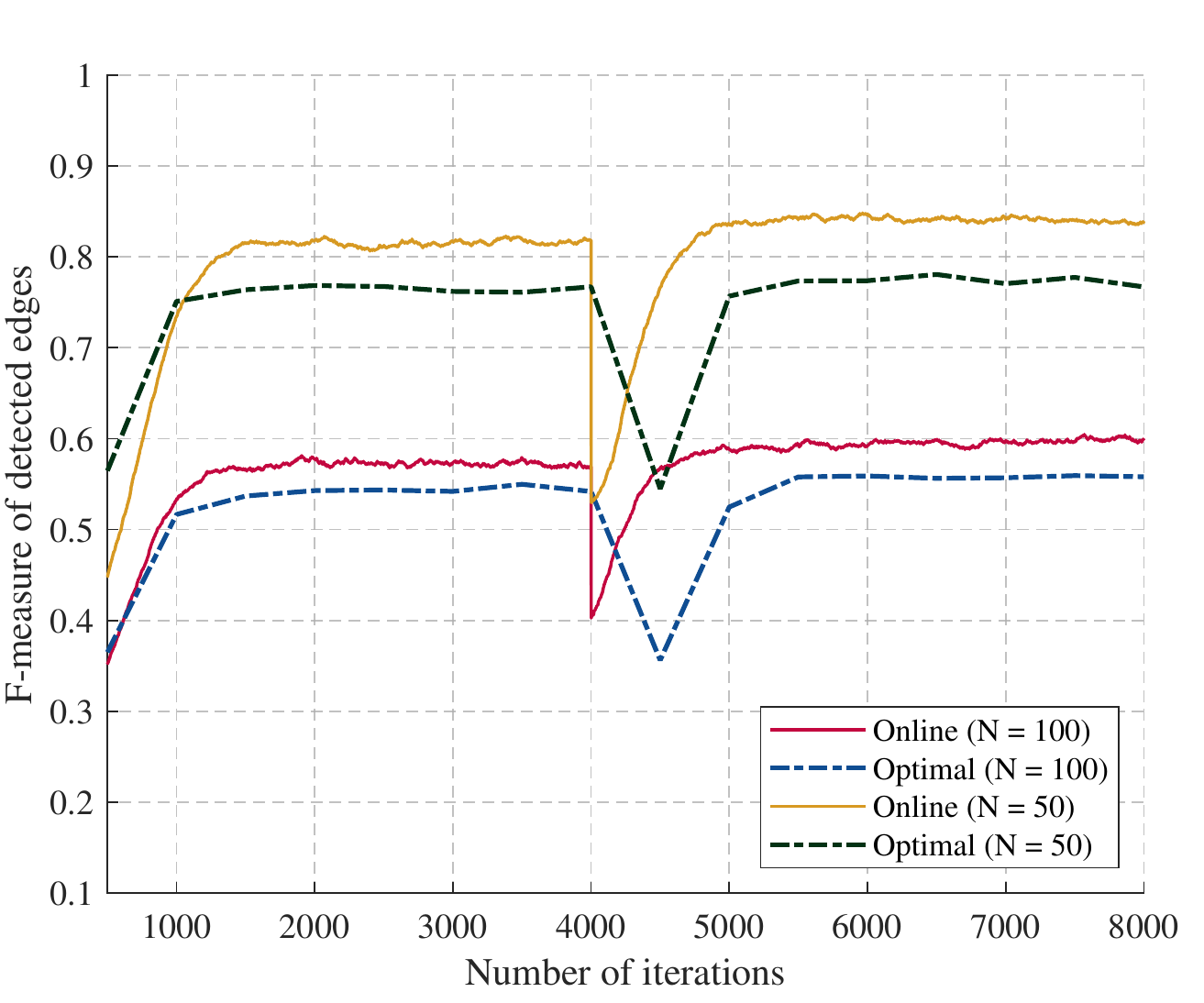}
    %\end{minipage}}
    %\vspace{-0.1in}
    \caption{Mean of objective value (8) (top), and F-measure of detected edges (bottom) as a function of acquired time samples (iteration). These results indicate that Algorithm \ref{A:online} can effectively track its offline counterpart. Minor F-measure gaps are due to thresholding of edge weights.}
    \vspace{-0.2in}
    \label{fig3}
\end{figure}

Fig.~\ref{fig3} (top) shows that after around $1000$ time samples (iterations) the objective value of the online Algorithm \ref{A:online} reaches the optimal value of its time-varying batch counterpart. The observed oscillation depends on the step size $\mu_t$ in Algorithm~\ref{A:online}. As expected, increasing the step size leads to faster convergence with larger oscillations. Conversely, reducing the step size leads to smoother and slower convergence. The other noteworthy observation is that the objective value markedly increases when the graph changes (after $4000$ time samples), but the online algorithm can effectively track the dynamic graph after a sufficient number of samples have been acquired. A similar trend can be observed for the F-measure of the detected edges (defined as the harmonic mean between edge precision and recall); see Fig.~\ref{fig3} (bottom).\vspace{2pt}

\noindent \textbf{Financial market.} Here, we perform a study involving real financial data. Since there is no ground-truth dynamic network, we endeavor to indirectly validate the proposed method by commenting on the intuitive structure observed from the learned sequence of graphs. We consider the daily stock prices for ten large American companies, including e.g., Microsoft (MSFT), Apple (AAPL) and Amazon (AMZN). %, Facebook (FB), Alphabet class A shares (GOOGL), Alphabet class C shares (GOOG), Johnson \& Johnson (JNJ), Berkshire Hathaway (BRK-B), Visa (V), and Procter \& Gamble (PG). 
We collect their daily stock prices from Yahoo! Finance over the time period from May 1st, 2019 to August 1st, 2020, %. The aforementioned time period consists of $317$ days worth of data, and it 
which overlaps with the very recent COVID-19 pandemic that led to significant market instabilities. Under normal circumstances, we would expect limited variations in the network describing the pairwise relationships between the chosen stock prices, since these large companies are well-established in the market~\cite{hallac17}. However, events like COVID-19 can cause abrupt changes in said network. We run Algorithm~\ref{A:online} to estimate daily graphs in order to monitor the sudden changes in the stock market. Following studies like~\cite{hallac17,cardoso20}, we quantify the variation of the network via relative temporal deviation $\| \bbW_t - \bbW_{t-1}\|_{F} / \| \bbW_{t-1}\|_{F}$. We also learn networks from signals given by the relative temporal variation of stock prices.
% in this test case, /// where the underlying graph is not readily available

\begin{figure}[t]
    \centering
   % \centerline{\begin{minipage}[c]{\linewidth}
    \centering
    \includegraphics[width=\linewidth]{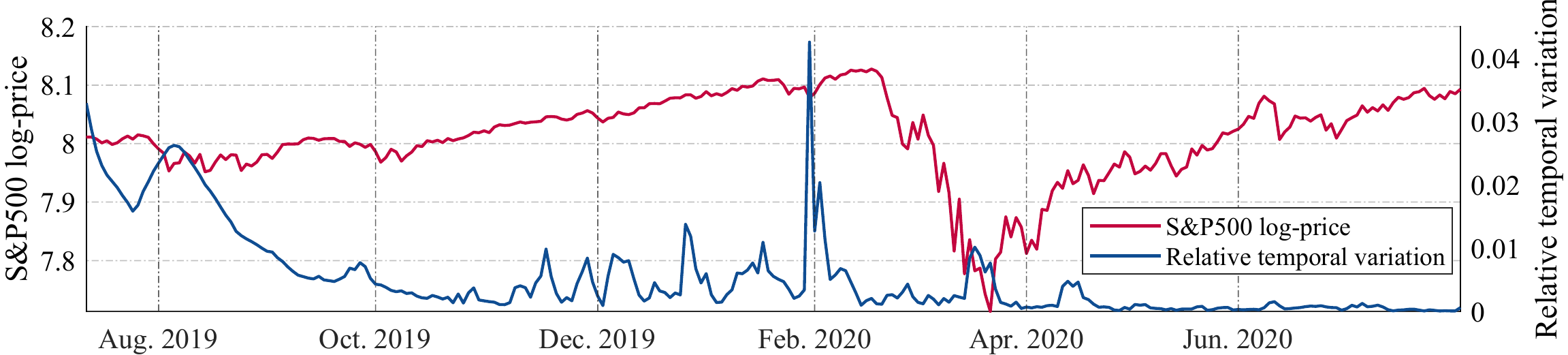}
   % \end{minipage}}
    %\vspace{-0.05in}
    \caption{The S\&P500 log-price per day (red). The daily relative temporal variation of the learned graphs (blue). The temporal variation indicates some sudden changes which can be due to e.g., COVID-induced market tensions.}
    \vspace{-0.3in}
    \label{fig8}
\end{figure}

In Fig.~\ref{fig8} (red) we plot the S\&P500 log-price as an indicator of the market's condition. The relative temporal variation of the learned graphs is illustrated in Fig.~\ref{fig8} (blue), which are obtained by setting the regularization parameters as  $\alpha=0.316$, and $\beta=0.05$. Fig.~\ref{fig8} (red) shows that the COVID-19 impacts on the markets started in February 2020 and it got to its worse situation during March 2020. The following sudden changes are apparent by inspection of the spikes in Fig.~\ref{fig8} (blue): (i)~Sep'19, (ii)~Nov'19, (iii)~Dec'19, (iv)~Jan'20, (v)~Mar'20, and (vi)~Apr'20. These abrupt changes are consistent with major events occurring during these time periods. In Sep'19 there was a congressional hearing regarding the impeachment inquiry of President Trump. The changes in Nov'19 can be explained by the optimism surrounding the U.S.-China trade negotiations. The U.S. House of Representatives' voted to impeach President Trump and that caused another sudden change in Dec'19. The big spike at the end of Jan'20 is probably the result of the World Health Organization (WHO) declaring a global health emergency due to the COVID-19 pandemic. In the middle of Mar'20, President Trump declared a national emergency due to COVID-19. The epidemic appears to continue (adversely) affecting the market well into Apr'20, when the daily case counts reached record values of $30000$.

\begin{figure}[t]
    \centering
    %\centerline{\begin{minipage}[c]{\linewidth}
    \centering
    \includegraphics[width=0.85\linewidth]{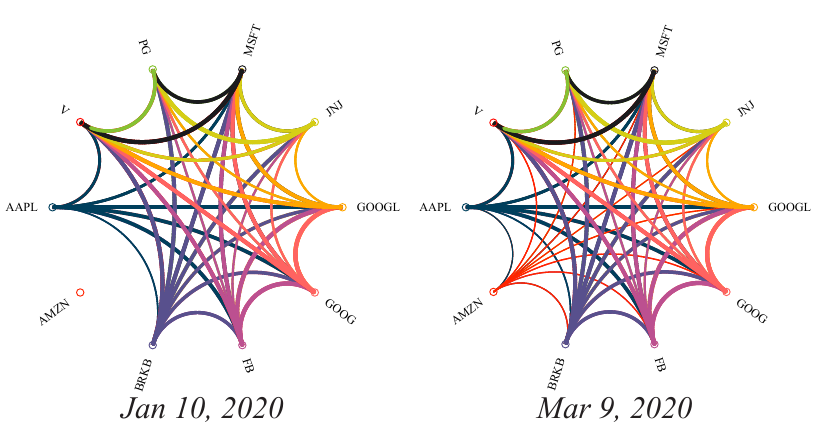}
    %\end{minipage}}
    %\vspace{-0.1in}
    \caption{The estimated network of the market over two different days: January 10th, 2020, and March 9th, 2020. Graph connectivity increases at the time of crisis when all the stock prices usually drop (right). While in stable times (left), the graph tends to be more loosely connected.}
    \vspace{-0.3in}
    \label{fig:market}
\end{figure}
%
% \begin{figure*}
%     \centering
%     \begin{minipage}{0.6\textwidth}
%     \centering
%     \includegraphics[width=\linewidth]{figures/fig_gspc_td_eusipco.pdf}
%     \end{minipage}
%     \begin{minipage}{0.35\textwidth}
%     \centering
%     \includegraphics[width=\linewidth]{figures/fig5_graph.pdf}
%     \end{minipage}
%     \caption{Caption}
%     \label{fig:my_label}
% \end{figure*}

We also learned networks from graph signals representing the relative temporal variation (single-day discrete gradients) of the stock prices, instead of the raw daily values considered so far. We set $\alpha=3.162$, and $\beta=0.088$. Two instances of the learned graphs (Jan. 10, 2020, and Mar. 9, 2020) are depicted in Fig.~\ref{fig:market}. The stock prices dropped sharply during the economic crisis in Mar'20; see Fig.~\ref{fig8} (red). This has an equalization effect on the stock price gradients, which suddenly become negative for all nodes (companies). Therefore, the Mar. 9 graph becomes more connected as a consequence of the imposed smoothness prior. However, during relatively normal operation of the market (e.g, during early Jan'20) the temporal stock-price gradients will be less-tightly coupled (possibly fluctuating up or down depending on some other latent effects). Accordingly, the learned graph tends to be relatively more disconnected as seen in Fig.~\ref{fig:market} (left).

%%%%%%%%%%%%%%%%%%%%%%%%%%%%%%%%%%%%%%%%%%%%%%%%%%%%%%%%%%%%%%%%%%%%%

\section{Conclusion}\label{S:conclusions}

%%%%%%%%%%%%%%%%%%%%%%%%%%%%%%%%%%%%%%%%%%%%%%%%%%%%%%%%%%%%%%%%%%%%%

In this paper, we developed a novel online algorithm to learn graphs from streaming smooth signals. We exploited recent advances in time-varying convex optimization to derive efficient iterations that are guaranteed to track the (slowly) time-varying optimal solution of the batch estimator. The tracking ability of the proposed method is corroborated using synthetic and real-world financial market data.

%%%%%%%%%%%%%%%%%%%%%%%%%%%%%%%%%%%%%%%%%%%%%%%%%%%%%%%%%%%%%%%%%%%%%

\bibliographystyle{IEEEtran}
\bibliography{refs.bib}

\begin{thebibliography}{10}
\providecommand{\url}[1]{#1}
\csname url@samestyle\endcsname
\providecommand{\newblock}{\relax}
\providecommand{\bibinfo}[2]{#2}
\providecommand{\BIBentrySTDinterwordspacing}{\spaceskip=0pt\relax}
\providecommand{\BIBentryALTinterwordstretchfactor}{4}
\providecommand{\BIBentryALTinterwordspacing}{\spaceskip=\fontdimen2\font plus
\BIBentryALTinterwordstretchfactor\fontdimen3\font minus
  \fontdimen4\font\relax}
\providecommand{\BIBforeignlanguage}[2]{{%
\expandafter\ifx\csname l@#1\endcsname\relax
\typeout{** WARNING: IEEEtran.bst: No hyphenation pattern has been}%
\typeout{** loaded for the language `#1'. Using the pattern for}%
\typeout{** the default language instead.}%
\else
\language=\csname l@#1\endcsname
\fi
#2}}
\providecommand{\BIBdecl}{\relax}
\BIBdecl

\bibitem{ortega18}
A.~{Ortega}, P.~{Frossard}, J.~Kova\u{c}evi\'{c}, J.~M.~F. {Moura}, and
  P.~{Vandergheynst}, ``Graph signal processing: Overview, challenges, and
  applications,'' \emph{Proc. IEEE}, vol. 106, no.~5, pp. 808--828, 2018.

\bibitem{kolaczyk09}
E.~D. Kolaczyk, \emph{Statistical Analysis of Network Data: Methods and
  Models}.\hskip 1em plus 0.5em minus 0.4em\relax New York, NY:
  Springer\hyp{}Verlag, 2009.

\bibitem{mateos19}
G.~{Mateos}, S.~{Segarra}, A.~G. {Marques}, and A.~{Ribeiro}, ``Connecting the
  dots: Identifying network structure via graph signal processing,'' \emph{IEEE
  Signal Process. Mag.}, vol.~36, no.~3, pp. 16--43, 2019.

\bibitem{dong2019learning}
X.~{Dong}, D.~{Thanou}, M.~{Rabbat}, and P.~{Frossard}, ``Learning graphs from
  data: A signal representation perspective,'' \emph{IEEE Signal Process.
  Mag.}, vol.~36, no.~3, pp. 44--63, 2019.

\bibitem{giannakis18}
G.~B. Giannakis, Y.~Shen, and G.~V. Karanikolas, ``Topology identification and
  learning over graphs: Accounting for nonlinearities and dynamics,''
  \emph{Proc. IEEE}, vol. 106, no.~5, pp. 787--807, 2018.

\bibitem{segarra2016topoidTSP16}
S.~Segarra, A.~Marques, G.~Mateos, and A.~Ribeiro, ``Network topology inference
  from spectral templates,'' \emph{IEEE Trans. Signal Inf. Process. Netw.},
  vol.~3, no.~3, pp. 467--483, Aug. 2017.

\bibitem{rasoul20}
R.~Shafipour and G.~Mateos, ``Online topology inference from streaming
  stationary graph signals with partial connectivity information,''
  \emph{Algorithms}, vol.~13, no.~9, pp. 1--19, Sep. 2020.

\bibitem{kalofolias16}
V.~Kalofolias, ``How to learn a graph from smooth signals,'' in \emph{Artif.
  Intel. and Stat. (AISTATS)}, 2016, pp. 920--929.

\bibitem{dong16}
X.~{Dong}, D.~{Thanou}, P.~{Frossard}, and P.~{Vandergheynst}, ``Learning
  {L}aplacian matrix in smooth graph signal representations,'' \emph{IEEE
  Trans. Signal Process.}, vol.~64, no.~23, pp. 6160--6173, 2016.

\bibitem{kalofolias2019iclr}
V.~Kalofolias and N.~Perraudin, ``Large scale graph learning from smooth
  signals,'' in \emph{Int. Conf. Learning Representations (ICLR)}, 2019.

\bibitem{chepuri17}
S.~P. {Chepuri}, S.~{Liu}, G.~{Leus}, and A.~O. {Hero}, ``Learning sparse
  graphs under smoothness prior,'' in \emph{IEEE Intl. Conf. Acoust., Speech
  and Signal Process. (ICASSP)}, 2017, pp. 6508--6512.

\bibitem{kalofolias17}
V.~{Kalofolias}, A.~{Loukas}, D.~{Thanou}, and P.~{Frossard}, ``Learning time
  varying graphs,'' in \emph{IEEE Intl. Conf. Acoust., Speech and Signal
  Process. (ICASSP)}, 2017, pp. 2826--2830.

\bibitem{cardoso20}
J.~V. d.~M. Cardoso and D.~P. Palomar, ``Learning undirected graphs in
  financial markets,'' \emph{arXiv preprint arXiv:2005.09958}, 2020.

\bibitem{koki20}
K.~Yamada, Y.~Tanaka, and A.~Ortega, ``Time-varying graph learning with
  constraints on graph temporal variation,'' \emph{arXiv preprint
  arXiv:2001.03346 [eess.SP]}, 2020.

\bibitem{madden19}
L.~{Madden}, S.~{Becker}, and E.~{Dall’Anese}, ``Online sparse subspace
  clustering,'' in \emph{IEEE Data Sci. Wrksp.}, 2019, pp. 248--252.

\bibitem{vlaski2018online}
S.~{Vlaski}, H.~P. {Mareti{\'c}}, R.~{Nassif}, P.~{Frossard}, and A.~H.
  {Sayed}, ``Online graph learning from sequential data,'' in \emph{IEEE Data
  Sci. Wrksp.}, 2018.

\bibitem{marques20}
A.~G. Marques, S.~Segarra, and G.~Mateos, ``Signal processing on directed
  graphs,'' \emph{arXiv preprint arXiv:2008.00586 [eess.SP]}, 2020.

\bibitem{zhou04}
D.~Zhou and B.~Sch{\"o}lkopf, ``A regularization framework for learning from
  graph data,'' in \emph{Int. Conf. Mach. Learning (ICML)}, 2004.

\bibitem{friedman01}
T.~Hastie, R.~Tibshirani, and J.~Friedman, \emph{The Elements of Statistical
  Learning}, 2nd~ed.\hskip 1em plus 0.5em minus 0.4em\relax New York: Springer,
  2009.

\bibitem{boyd14}
N.~Parikh and S.~Boyd, ``Proximal algorithms,'' \emph{Foundations and Trends in
  optimization}, vol.~1, no.~3, p. 127–239, 2014.

\bibitem{saboksayr20}
S.~S. Saboksayr, G.~Mateos, and M.~Cetin, ``Online discriminative graph
  learning from multi-class smooth signals,'' \emph{arXiv preprint
  arXiv:2101.00184 [eess.SP]}, 2021.

\bibitem{beck18}
A.~Beck, \emph{First\hyp{}order Methods in Optimization}.\hskip 1em plus 0.5em
  minus 0.4em\relax Philadelphia, PA: Society for Industrial and Applied
  Mathematics, 2018.

\bibitem{hallac17}
D.~Hallac, Y.~Park, S.~Boyd, and J.~Leskovec, ``Network inference via the
  time-varying {Graphical Lasso},'' in \emph{Proc. of the 23rd ACM SIGKDD
  International Conference on Knowledge Discovery and Data Mining}, 2017, pp.
  205--213.

\end{thebibliography}

%%%%%%%%%%%%%%%%%%%%%%%%%%%%%%%%%%%%%%%%%%%%%%%%%%%%%%%%%%%%%%%%%%%%%

\end{document}